\documentclass[conference]{IEEEtran}
\IEEEoverridecommandlockouts
\usepackage{cite}
\usepackage{amsmath,amssymb,amsfonts}
\usepackage{algorithmic}
\usepackage{graphicx}
\usepackage{textcomp}
\usepackage{xcolor}

\DeclareMathAlphabet\mathbfcal{OMS}{cmsy}{b}{n}

\newcommand{\mat}[1]{\mathbf{#1}}

\def\BibTeX{{\rm B\kern-.05em{\sc i\kern-.025em b}\kern-.08em
    T\kern-.1667em\lower.7ex\hbox{E}\kern-.125emX}}
\begin{document}

\title{Detecting Gender Bias in Transformer-based Models: A Case Study on BERT\\
\vspace{-15pt}
}


\author{
\IEEEauthorblockN{ Bingbing Li\textsuperscript{[1]}, Hongwu Peng\textsuperscript{[1]}, Rajat Sainju\textsuperscript{[1]}, Yueying Liang\textsuperscript{[1]}, Junhuan Yang\textsuperscript{[2]}, Lei Yang\textsuperscript{[2]},  \\ Weiwen Jiang\textsuperscript{[3]},  Binghui Wang\textsuperscript{[4]}, Hang Liu\textsuperscript{[5]}, and Caiwen Ding \textsuperscript{[1]}}
\IEEEauthorblockA{
\textsuperscript{[1]}University of Connecticut, Storrs, CT, USA. 
\textsuperscript{[2]}University of New Mexico, NM, USA
 \\
\textsuperscript{[3]}George Mason University, VA, USA. 
\textsuperscript{[4]}Illinois Institute of Technology, IL, USA.
\\
\textsuperscript{[5]}Stevens Institute of Technology, Hoboken, NJ, USA.
\\
\textsuperscript{[1]}\{bingbing.li, hongwu.peng, rajat.sainju, yueying.liang, caiwen.ding\}@uconn.edu,  \\
\textsuperscript{[2]}\{yangjh1993, leiyang\}@unm.edu,  
\textsuperscript{[3]}wjiang8@gmu.edu, 
\textsuperscript{[4]}bwang70@iit.edu,
\textsuperscript{[5]}hliu77@stevens.edu
\vspace{-15pt}
}
}

\maketitle

\begin{abstract}
In this paper, we propose a novel 
gender bias detection method by utilizing attention map for transformer-based models. We 1) give an intuitive gender bias judgement method by comparing the different relation degree between the genders and the occupation according to the attention scores, 2) design a gender bias detector by modifying the attention module, 3) insert the gender bias detector into different positions of the model to present the internal gender bias flow, and 4) draw the consistent gender bias conclusion by scanning the entire Wikipedia, a BERT pretraining dataset. We observe that 1) the attention matrices, $\mat{W_q}$ and $\mat{W_k}$ introduce much more gender bias than other modules (including the embedding layer) and  2) the bias degree changes periodically inside of the model (attention matrix $\mat{Q}$, $\mat{K}$, $\mat{V}$, and the remaining part of the attention layer (including the fully-connected layer, the residual connection, and the layer normalization module) enhance the gender bias while the averaged attentions reduces the bias). 
\end{abstract}

\begin{IEEEkeywords}
gender bias, transformer, attention, detection, analysis
\end{IEEEkeywords}

\section{Introduction}

Great success has been witnessed in computer vision (CV) and natural language processing (NLP), by utilizing attention-based transformer structure.
For instance, Transformer-based models have advanced the state-of-the-arts of image classification, object detection, and semantic segmentation in CV (e.g., ViT~\cite{dosovitskiy2020image}, DeiT~\cite{touvron2021training}, DETR~\cite{carion2020end}, Deformable DETR~\cite{zhu2020deformable}, Swin Transformer~\cite{liu2021swin}) and text classification, natural language inference, and question answering in NLP (e.g., BERT~\cite{devlin2018bert}, XLNet~\cite{yang2019xlnet}, RoBERTa~\cite{liu2019roberta}, MT-DNN~\cite{liu2019multi},  ALBERT~\cite{lan2019albert}, GPT v1-3~\cite{radford2018improving,radford2019language,brown2020language}, and T5~\cite{raffel2020exploring}). However, the inexplainability and 
bias introduced by the transformer-based models could 
become a main barrier for their real-world deployment
~\cite{mehrabi2021survey,fan2021interpretability}. 
Previous researches focus on the gender bias of the embedding layer or the output of the whole model~\cite{bolukbasi2016man,sheng2021societal}.
In this paper, we take the first step to study whether gender bias associated with occupations exists inside of the pretrained BERT. We observe that 1) the attention matrices, $\mat{W_q}$ and $\mat{W_k}$, introduce much more gender bias than other module (including the embedding layer) and  2) the bias degree changes periodically inside of the model ($\mat{Q}$, $\mat{K}$, $\mat{V}$, and the remaining part of the attention layer (including the fully-connected layer, the residual connection, and the layer normalization) enhance the gender bias while the averaged attentions reduce the bias).

\begin{figure}[t]%
\centering
 \includegraphics[width=3.3in]{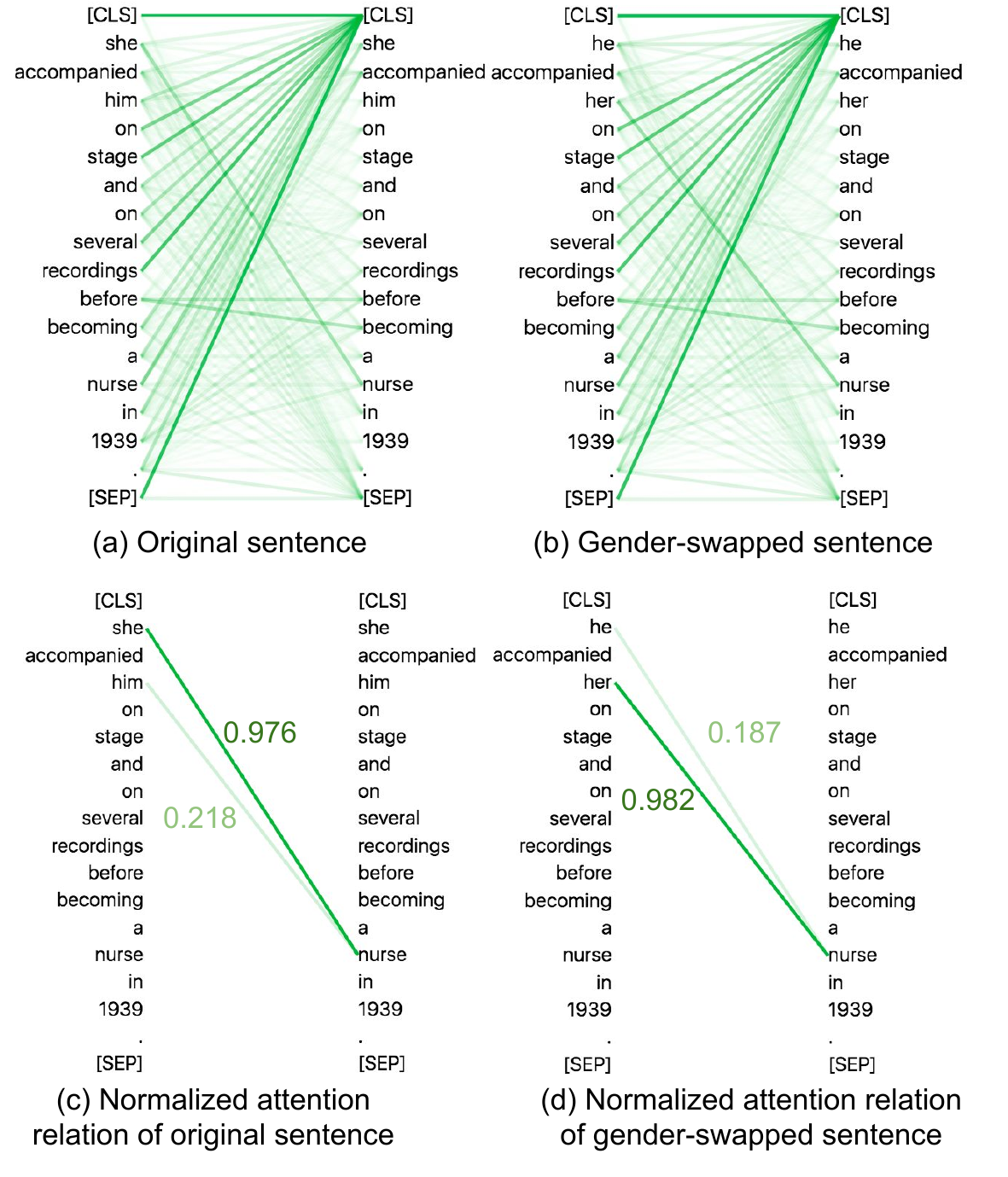}
\caption{To analysis the gender bias, we first swap the genders in the sentence to obtain the gender-swapped sentences (in (a) and (b)); and then obtain the corresponding attention connections between different gender pronouns and the occupation (in (c) and (d))}
\label{fig:attention_connection}%
\end{figure}


\begin{figure*}[!t]
\center
 \includegraphics[width=6in]{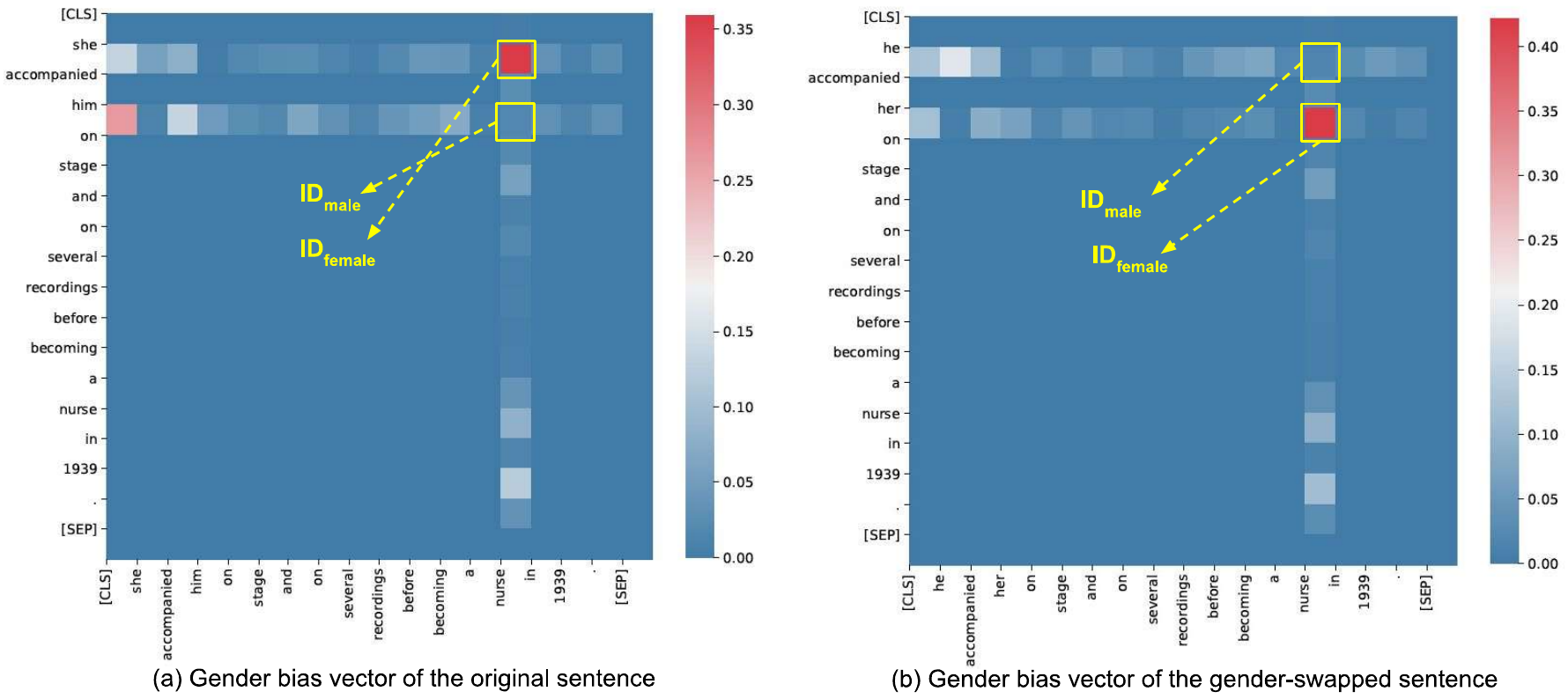} 
\caption{Attention score matrix extraction to derive the gender bias}
\label{fig:attention_connection_matrix}
\end{figure*}





\section{Attention module visualization}
Self-attention module plays an essential role in transformer-based language models, e.g. BERT. In this module, trainable matrices, $\mat{W_q}$, $\mat{W_k}$, and $\mat{W_v}$, are utilized to obtain attention matrices, $\mat{Q}$, $\mat{K}$, and $\mat{V}$, and then the attention score, $\mat{AS}$ and the averaged attentions, $\mat{AvgAttention}$, are derived as follows

\begin{eqnarray}
\small
\label{eq:vector_qkv}
\mat{Q} = \mat{input} * \mat{W_q}\\
\mat{K} = \mat{input} * \mat{W_k}\\
\mat{V} = \mat{input} * \mat{W_v}
\end{eqnarray}
\vspace{-0.25cm}
\begin{equation}
\small
\label{eq:attention}
\mat{AS}(\mat{Q}, \mat{K}) =  Softmax(\frac{\mat{Q} \times \mat{K}^{T}}{\sqrt{D_k}})
\end{equation}
\vspace{-0.25cm}
\begin{equation}
\small
\label{eq:attention}
\mat{AvgAttention}(\mat{Q}, \mat{K}, \mat{V}) =  \mat{AS}(\mat{Q}, \mat{K}) * \mat{V}
\end{equation}
where $\mat{input}$ is the output of the embedding layer for the first attention layer or the output of the previous attention layer for the remaining attention layers, $D_k$ represents the dimension of matrix $\mat{K}$. 

For each attention head (for BERT model, there are 12 heads for each layer and 12 layers, thus totally 144 heads), the dimension of the $\mat{AS}$ is $m*m$, which $m$ is the length of the input sentence and each element of the $\mat{AS}$ corresponds to the attention connection degree between two different words.
Fig. 1 (a) shows the attention connection between different words of the input sentence with the pretrained BERT model using bertviz toolbox~\cite{vig-2019-multiscale} and Fig. 1 (c) shows the attention connection between the occupation (e.g., "nurse") and the whole sentence.
We also extract the elements corresponding to different genders and the occupation for bias analysis as shown in 
Fig.~\ref{fig:attention_connection_matrix}.

\begin{figure}[htbp]%
\centering
 \includegraphics[width=2.8in]{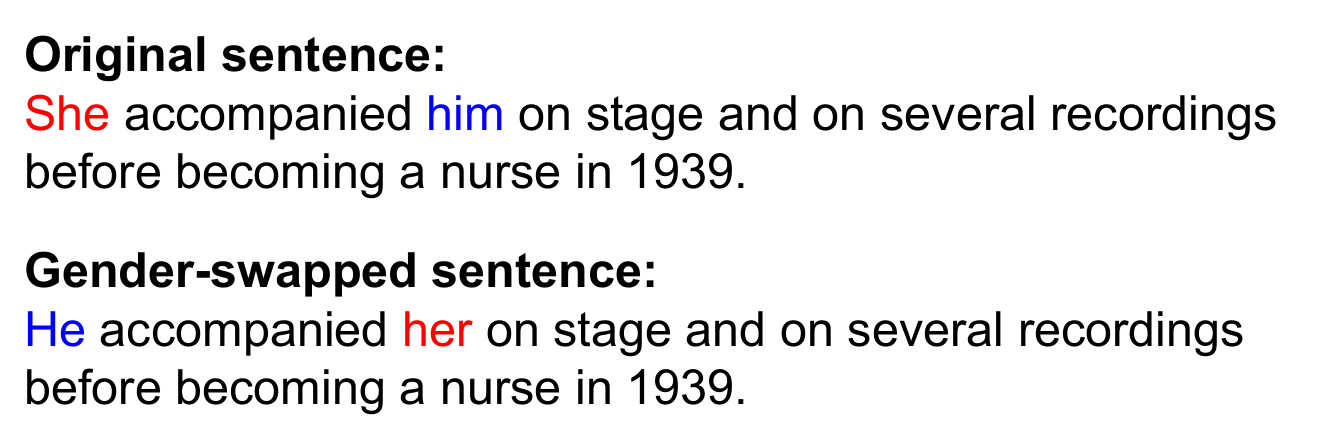}
\caption{Gender swapping}
\label{fig:sentence_swapped}%
\end{figure}

\section{Gender bias analysis based on the  attention map}

\subsection{Gender bias analysis based on gender-swapped sentences}
For each sentence, we 1) find the gender pronouns and the occupation (in our test, we choose the last occupation if there are more than one), 2) calculate the gender tendency value by summing all the $\mat{AS}$ elements associated with each gender, 3) swap the gender of the sentence according to~\cite{zhao2018gender,lu2020gender} to avoid the position effect on the gender bias judgement.

For input sentence $ST$, we obtain the index of the word associated with male and female, and calculate the gender tendency by summing all the $\mat{AS}$ elements associate with each gender as follows

\begin{eqnarray}
\small
\label{eq:gender_ID}
\left[i_1,i_2, \cdots, i_p\right] = {\rm ID}_{male}\left({\rm ST}\right))\\
\left[j_1,j_2, \cdots, j_q\right] = {\rm ID}_{female}\left({\rm ST}\right))\\
k = {\rm ID}_{occupation}\left({\rm ST}\right))
\end{eqnarray}

where $p$ and $q$ are the number of male and female pronouns in $ST$, $ID_{male}(*)$, $ID_{female}(*)$, and $ID_{occupation}(*)$ return the index of male and female pronouns and occupation, respectively. 

Then we extract the $\mat{AS}$ elements associated with genders as shown in Fig.~\ref{fig:attention_connection} and Fig. 2 (a). Specifically, we choose the attention score between the male pronoun (e.g., ``him'') and the occupation (e.g., ``nurse'') as the male tendency, $T_{male}$, and the attention score between the female pronoun (e.g., ``she'') and the occupation (e.g., ``nurse'') as the female tendency, $T_{female}$, as follows

\begin{eqnarray}
\small
\label{eq:gender_ID}
T_{male} = \sum\limits_{ \forall i \in {\rm ID}_{male} }{\rm AS}_{k,i}\\
T_{female} = \sum\limits_{ \forall j \in {\rm ID}_{female}}{\rm AS}_{k,j}
\end{eqnarray}

Then we derive the bias of the sentence, $bias_{ST}$ by normalizing [$T_{male}$, $T_{female}$] and calculate the difference as follows

\begin{eqnarray}
\small
\label{eq:gender_ID}
V_{male} = \frac{T_{male}}{\sqrt{T_{male}^2 + T_{female}^2 }}\\
V_{female} = \frac{T_{female}}{\sqrt{T_{male}^2 + T_{female}^2 }}\\
bias_{ST} = V_{male} - V_{female}
\end{eqnarray}

Finally, we swap the gender pronouns as shown in Fig.~\ref{fig:sentence_swapped}, obtain
the bias of the gender-swapped sentence, $bias_{ST_{swap}}$, and derive the final $degree_{biased}$ to determine the gender bias existence and degree (if $degree_{biased}$ is larger than 0, then gender bias is detected) as follows

\begin{eqnarray}
\small
\label{eq:gender_ID}
degree_{biased}=bias_{ST}*bias_{ST_{swap}}
\end{eqnarray}



\begin{figure}[!h]%
\centering
 \includegraphics[width=3.3in]{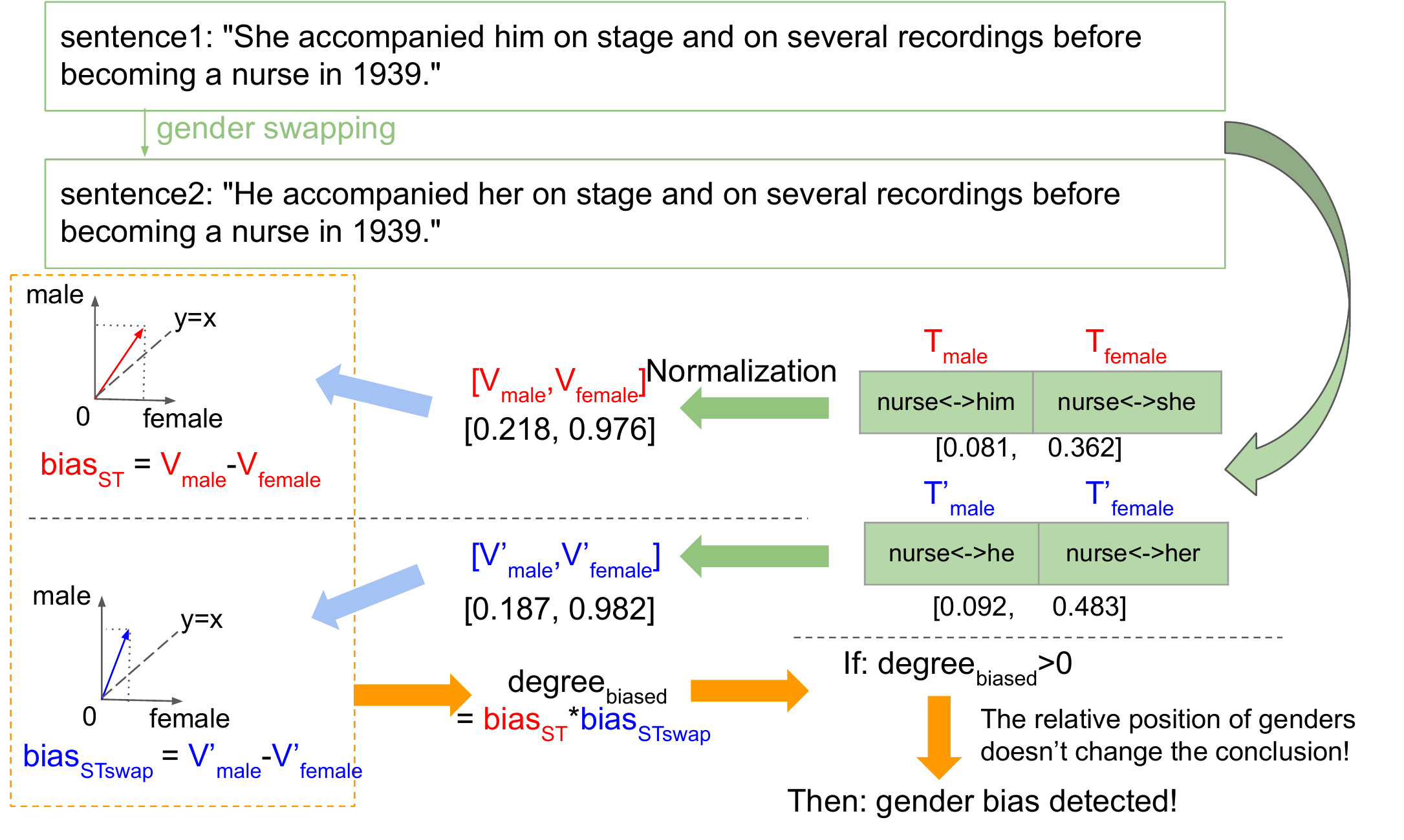}
\caption{Gender bias judgement method}
\label{fig:gender_bias_method}%
\end{figure}

The whole procedure is shown in Fig.~\ref{fig:gender_bias_method}.

\begin{figure*}[h!]%
\centering
 \includegraphics[width=0.9\linewidth]{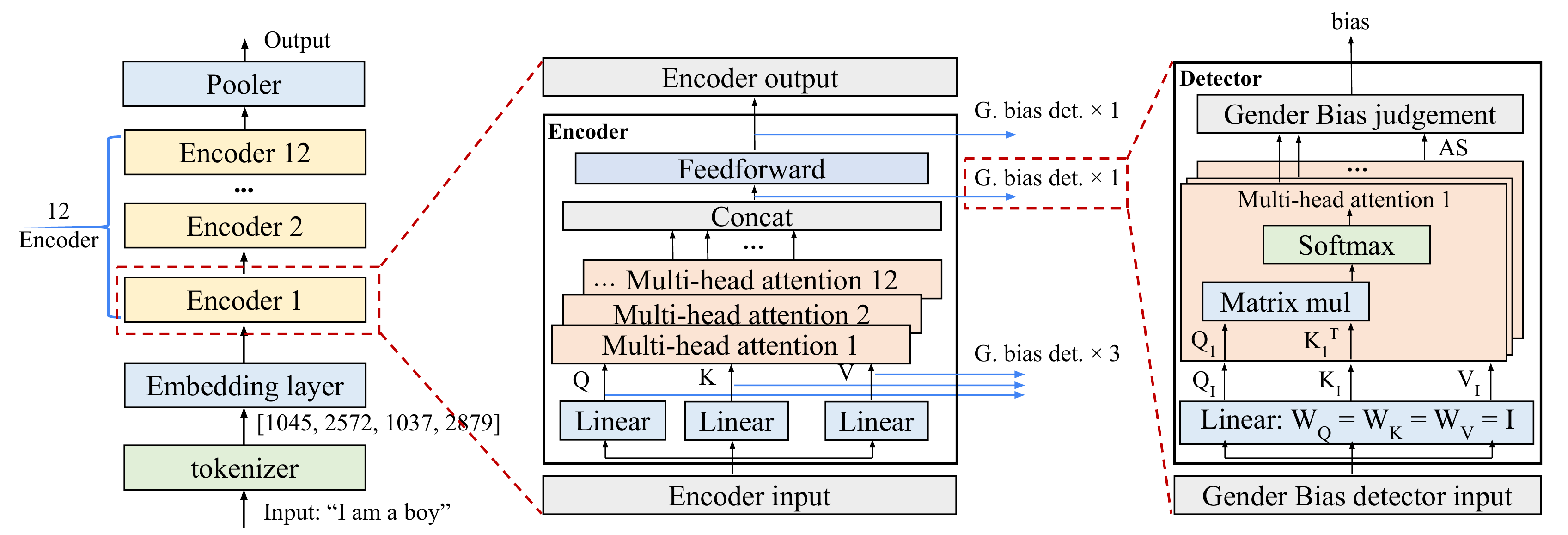}
\caption{Gender bias detectors in different position of the BERT model}
\label{fig:genderbias}%
\end{figure*}

\subsection{Gender bias detectors in different positions of the BERT model}
For the BERT model, the data flow inside of the BERT can be described as follows:
First, each word of the input sentence is converted into numbers using the tokenizer; then, word vectors are obtained after looking up the dictionary according to trained embedding layer matrix (by default, we use a 768-dimensional vector to represent each word) and we insert the first bias detector at the output of the embedding layer; then the output is connected to the attention layer (for BERT-base model, there are 12 attention layers) and we insert 3 bias detector to detect the bias of the 3 attention matrices, $\mat{Q}$, $\mat{K}$, and $\mat{V}$; then the averaged attention, $\mat{AvgAttention}$ are calculated, in which we insert the third bias detector; finally, we insert the forth bias detector at the output of the attention layer to detect the bias change between the averaged attention and the remaining part (including the fully-connected layer, the residual connection, and the layer normalization) of the attention layer, which we refer to collectively as residual attention part.

Fig.~\ref{fig:genderbias} shows the positions for gender bias detection inside of the BERT model and the structure of our bias detector. In the gender bias detector, we set the attention matrices, $\mat{W_q}$, $\mat{W_k}$, and $\mat{W_v}$, as identity matrices ($\mat{I}$) and use multi-head attention strategy to obtain attention scores and the gender bias judgement to detect the bias.


\section{Experiment}
\subsection{Dataset and sentences filtering}
\begin{figure}[htbp]%
\centering
 \includegraphics[width=3.3in]{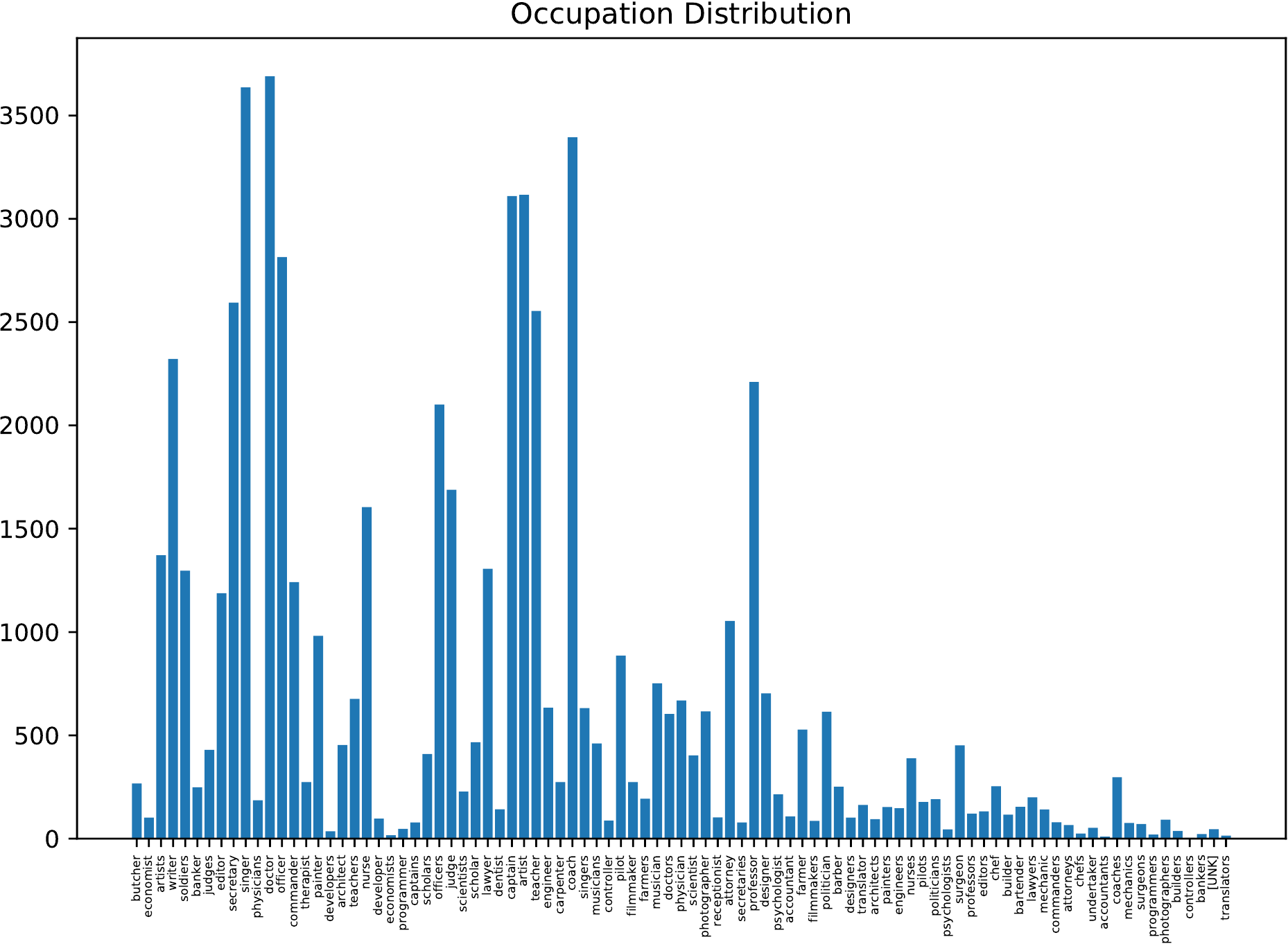}
\caption{Occupation distribution of our filtered dataset (totally 60,548 sentences) }
\label{fig:occupation_distribution}%
\end{figure}

We test our gender bias detection method on the BERT pretraining dataset, Wikipedia.
In this dataset, there are more than two million sentences. We extract the sentences by designing a filter that the expected sentence should include two opposite gender pronouns (e.g., "he" and "she") and one occupations (e.g., "nurse"). Finally,we obtain 60,548 sentences for gender bias detection.
Fig.~\ref{fig:occupation_distribution} shows the occupation distribution of the filtered dataset.

\begin{figure}[!b]%
\centering
 \includegraphics[width=3.5in]{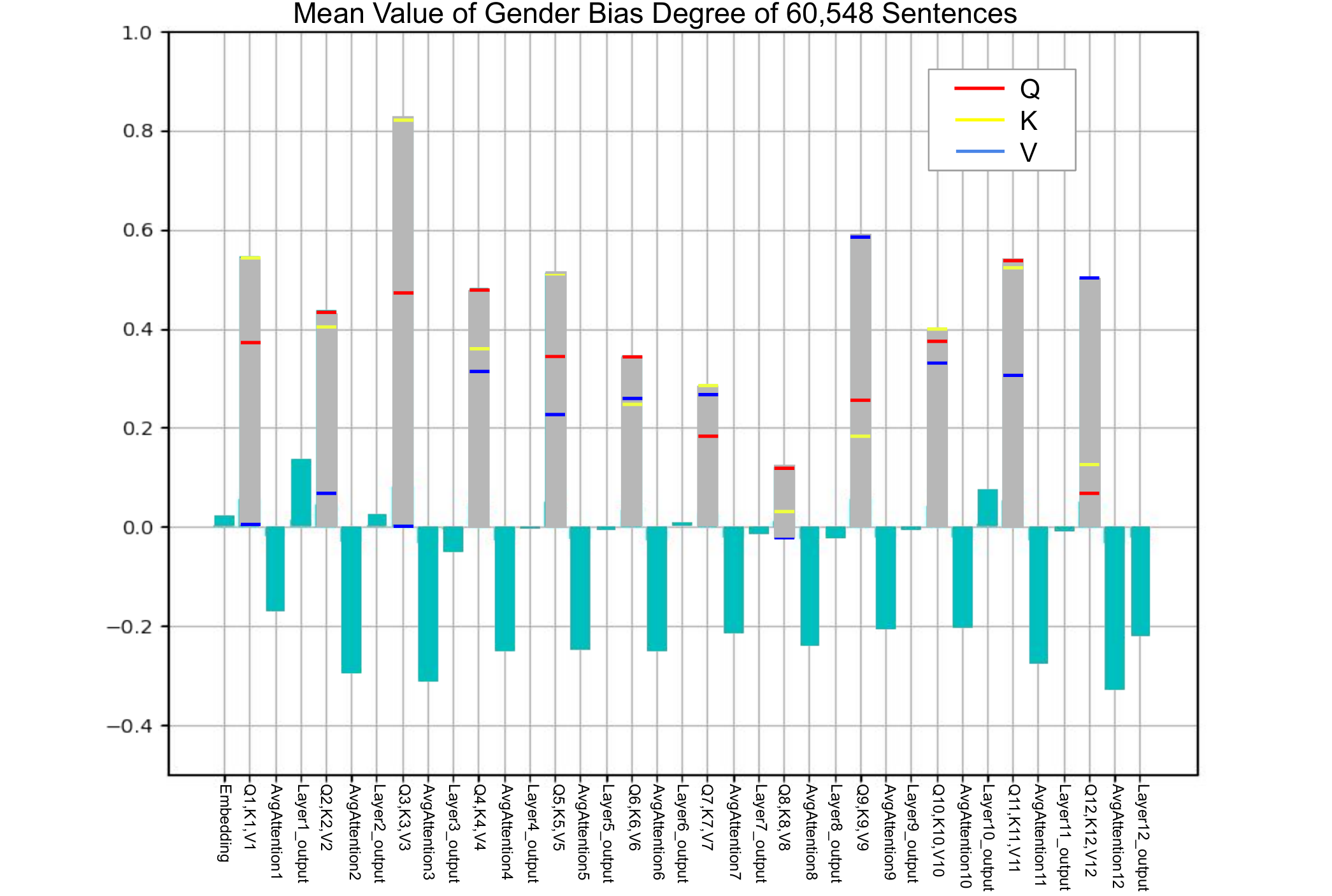}
\caption{Mean values of gender biases in different positions (embedding layer output, $\mat{Q}$, $\mat{K}$, $\mat{V}$, $\mat{AvgAttention}$, and attention layer output) inside of the BERT model for 60,548 test sentences. Curves between red dashed lines correspond to different detection positions in one attention layer.}
\label{fig:genderbias_prob}%
\end{figure}

\subsection{Detection results}
For each test sentence, we calculate the $degree_{biased}$ by swapping the gender pronouns of the input sentence and doing bias detection in different positions of the pretrained BERT model with the original and the gender-swapped sentences. Then we count the distribution characteristics among the whole test dataset of 60,548 sentences. 

Fig.~\ref{fig:genderbias_prob} shows the mean value $degree_{biased}$ across the whole test dataset in different positions as described in Fig.~\ref{fig:genderbias}. Curves between green dashed lines correspond to different detection positions in one attention layer. We plot the biases introduced by $\mat{W_q}$, $\mat{W_k}$, $\mat{W_v}$ separately since they work in parallel. From the results, we observe that 1) $\mat{W_q}$ and $\mat{W_k}$ show larger bias tendency than other positions while $W_v$ introduce much smaller bias; 2) the averaged attention, $\mat{AvgAttention}$, have negative bias values and thus do not show any bias tendency; 3) the layer outputs show larger bias tendency and help us to analysis the bias enhancement of the remaining part of the attention layer.

\begin{figure}[!t]%
\centering
 \includegraphics[width=3in]{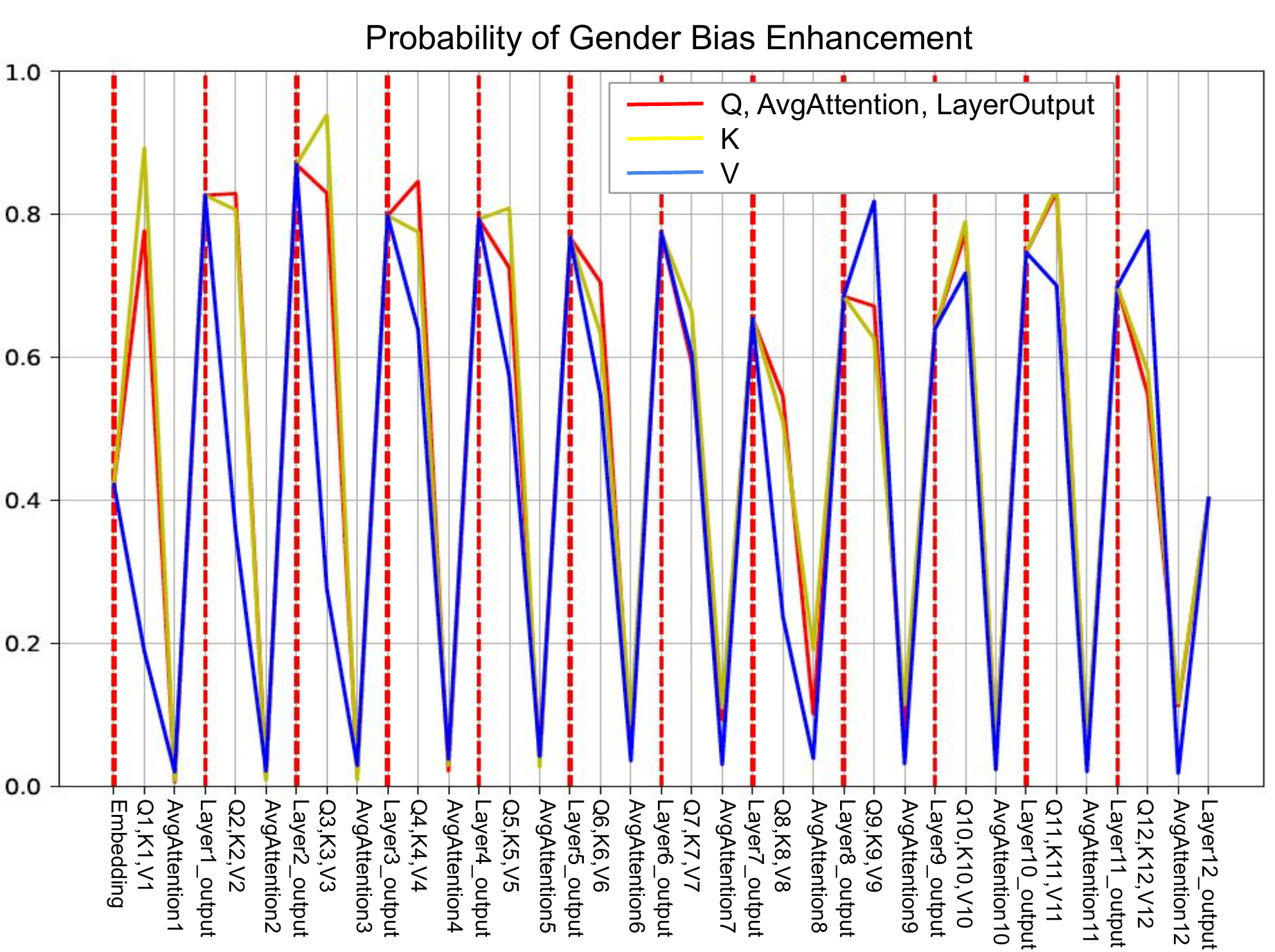}
\caption{Probability of gender bias enhancement between adjacent detection positions. Curves between red dashed lines correspond to different detection positions (embedding layer output, $\mat{Q}$, $\mat{K}$, $\mat{V}$, $\mat{AvgAttention}$, and attention layer output) in one attention layer.}
\label{fig:genderbias_trend}%
\end{figure}

\begin{figure}[!b]%
\centering
 \includegraphics[width=3.0in]{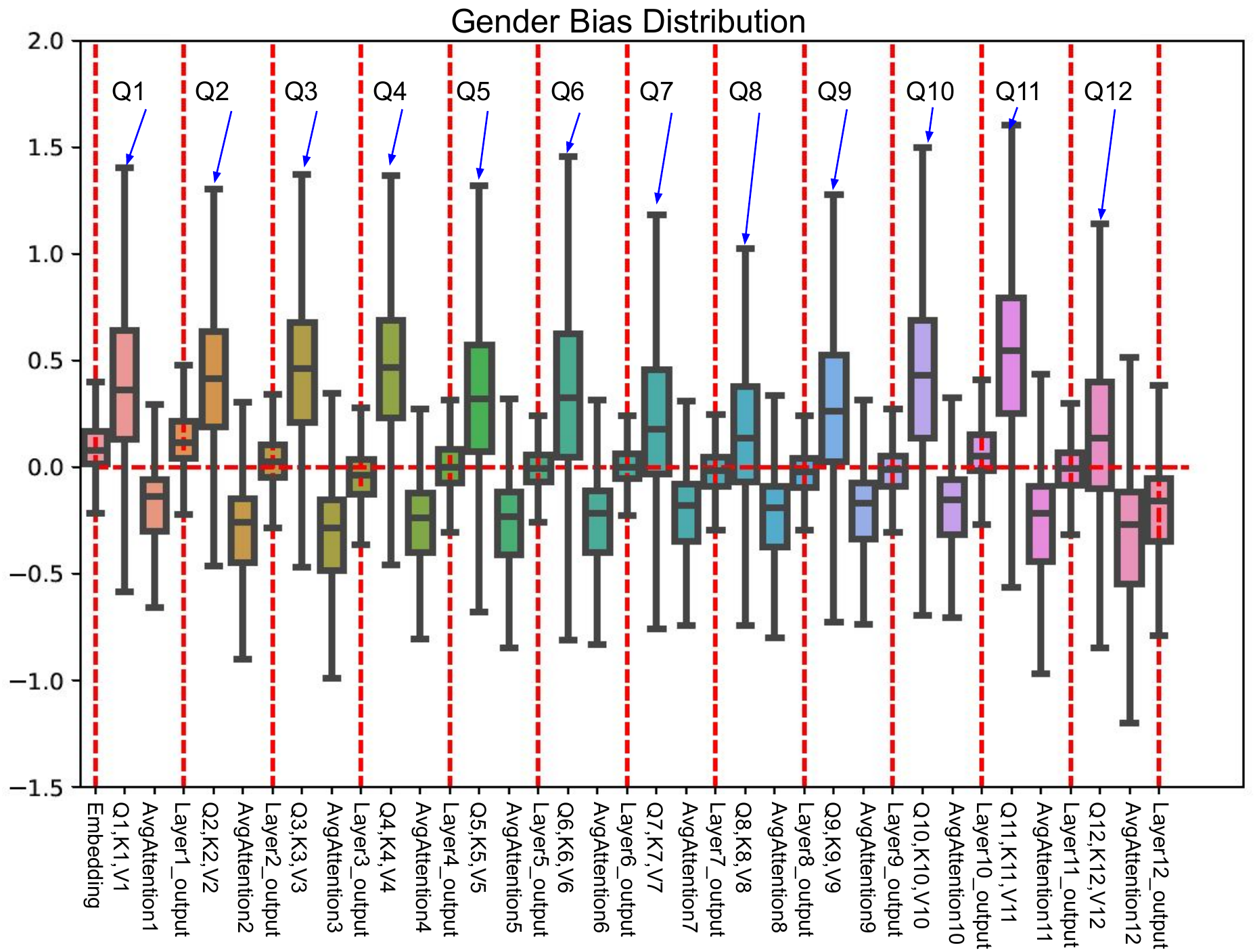}
\caption{Gender bias degree distribution in different positions of the pretrained BERT model (embedding layer output, $\mat{Q}$, $\mat{AvgAttention}$, and attention layer output). Curves between red dashed lines correspond to different detection positions in one attention layer.}
\label{fig:genderbias_distribute_wq}%
\end{figure}

\begin{figure}[!t]%
\centering
 \includegraphics[width=3.0in]{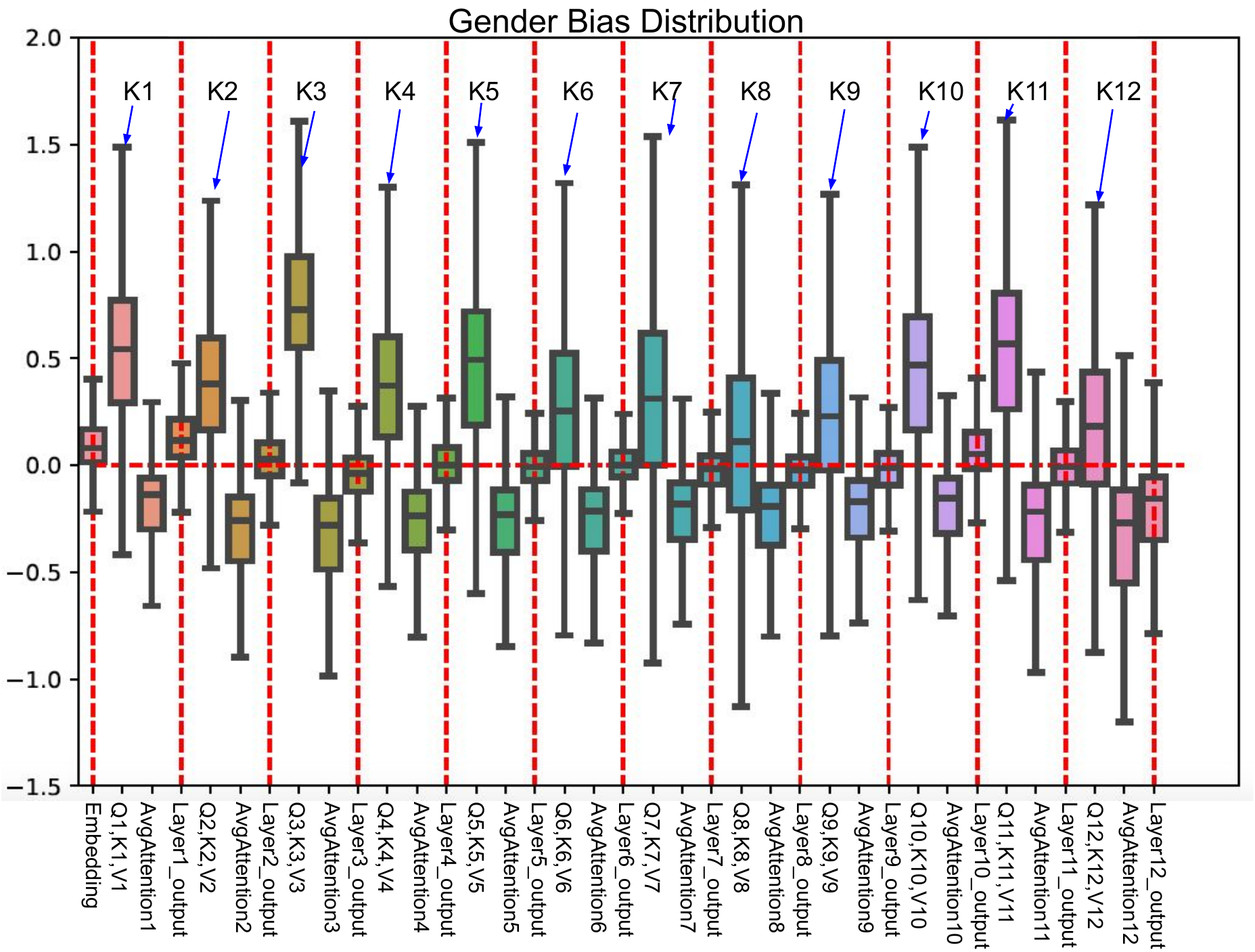}
\caption{Gender bias degree distribution in different positions of the pretrained BERT model (embedding layer output, $\mat{K}$, $\mat{AvgAttention}$, and attention layer output). Curves between red dashed lines correspond to different detection positions in one attention layer.}
\label{fig:genderbias_distribute_wk}%
\end{figure}

\begin{figure}[htbp]%
\centering
 \includegraphics[width=3.0in]{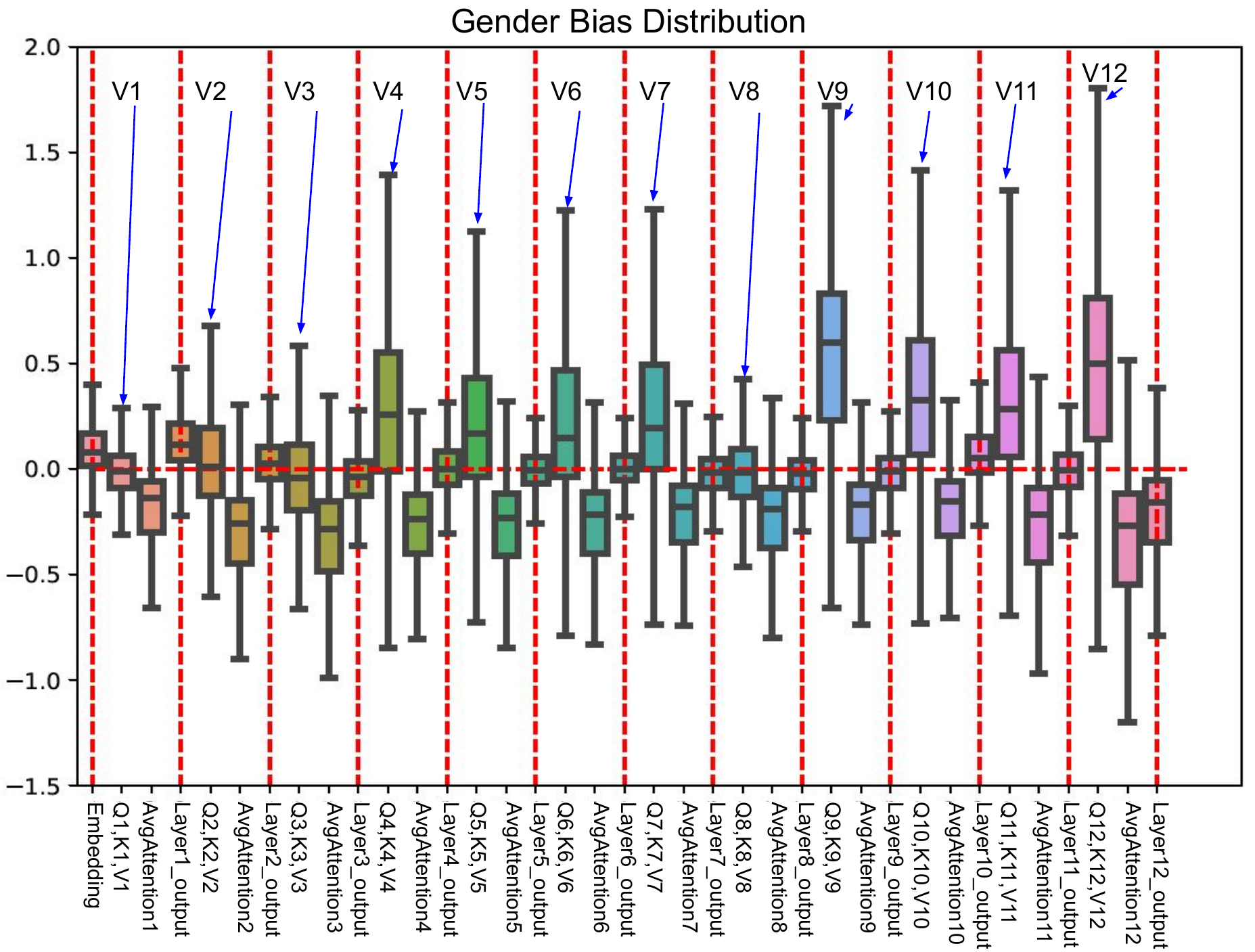}
\caption{Gender bias degree distribution in different positions of the pretrained BERT model (embedding layer output, $\mat{V}$, $\mat{AvgAttention}$, and attention layer output). Curves between red dashed lines correspond to different detection positions in one attention layer.}
\label{fig:genderbias_distribute_wv}%
\end{figure}

\begin{figure}[htbp]%
\centering
 \includegraphics[width=3.5in]{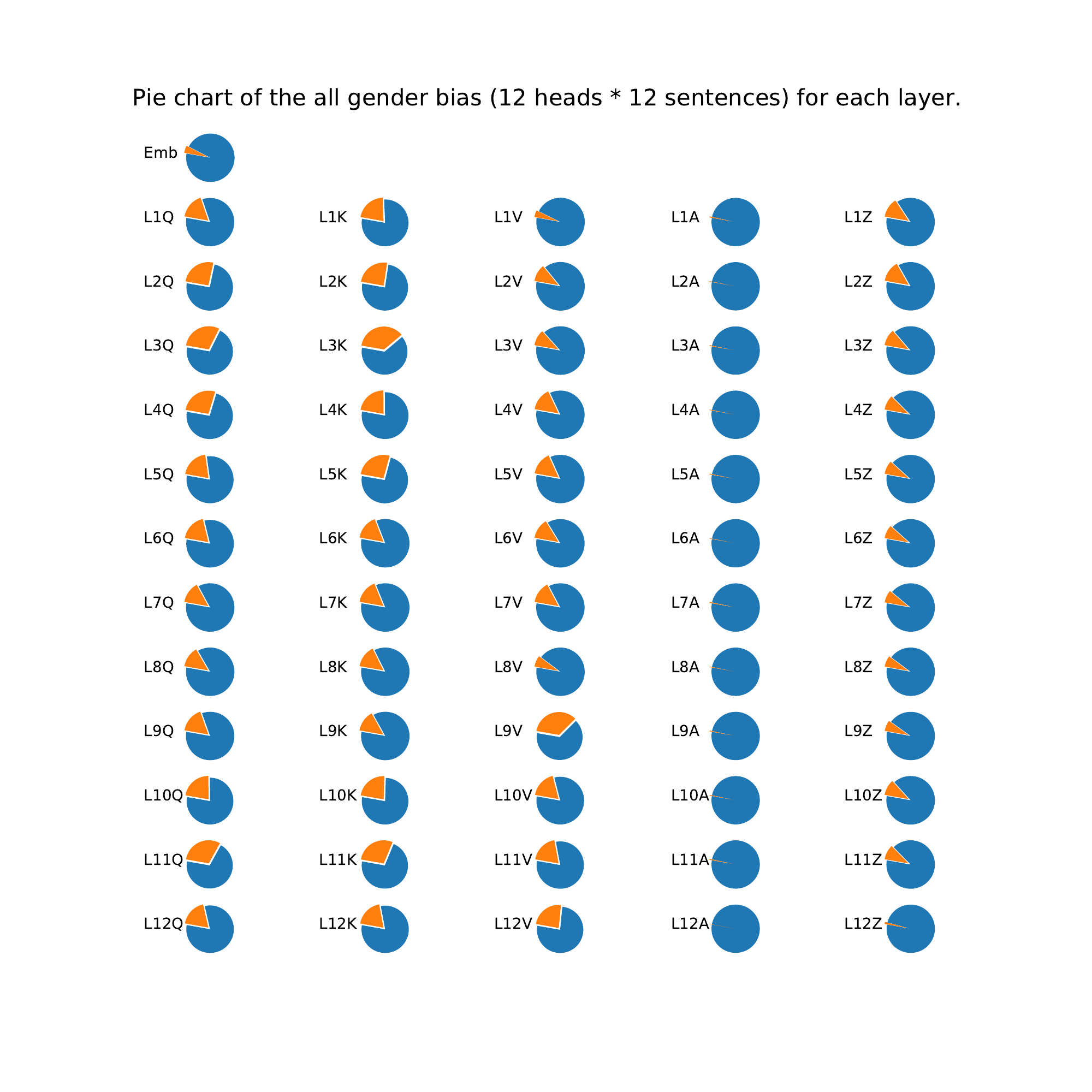}
\caption{Percentage of unbiased (in blue color) and biased (in orange color) heads for 60,548 sentences. Emb: Embedding output. LiQ/Lik/LiV/LiA/LiZ: $\mat{Q}$, $\mat{K}$, $\mat{V}$, $\mat{AvgAttention}$, final output of the $i_{th}$ layer.}
\label{fig:genderbias_positive}%
\end{figure}

To make the relative gender bias change more clearly between adjacent detection positions to check the bias enhancement, we calculate the difference between adjacent detection positions and also the percentage of bias enhancement (the bias of the current position is larger than the previous one) as shown in Fig.~\ref{fig:genderbias_trend}. We conclude that 1) attention matrices, $\mat{W_q}$, $\mat{W_k}$, and $\mat{Wv}$, have higher probability than other modules to enhance the bias; 2) averaged attention usually does not enhance the bias; and 3) the remaining part of the attention layer enhance the bias again.

Additionally, we show the gender bias distribution in different positions inside of the model. We use box plot to show the gender bias value distributions as Fig.~\ref{fig:genderbias_distribute_wq} to Fig.~\ref{fig:genderbias_distribute_wv}. We observe that 1) $\mat{Q}$, $\mat{K}$, and $\mat{V}$ show significant bias at each layer; 2) $\mat{Q}$ and $\mat{K}$ have larger bias than other positions of the model, including $\mat{W_v}$.

Furthermore, we show the percentage of the biased head (corresponding to the positive $degree_{biased}$) in Fig.~\ref{fig:genderbias_positive}. We conclude that 1) $\mat{W_q}$ and $\mat{W_k}$ increase the percentage of biased heads, which is much larger than the percentage of bias heads introduced by other modules; 2) $\mat{W_v}$ leads to fewer biased heads and the residual part of the attention layer increase the percentage of the biased heads. This coincides the distribution of the mean value of bias in different position of the BERT model.


\section{Conclusion}
In this paper, we propose a novel gender bias detection method based on attention map for transformer-based models. We extract the attention scores of the corresponding gender pronouns and occupation, swap the gender pronouns to avoid position effect on bias judgement, and check the consistency of the gender bias associated with the occupation. The gender bias distribution conclusions are drawn by scanning the whole filter dataset obtained from Wikipedia, a BERT pretraining dataset. 
We take the first attempt to study the gender bias inside of the transformer-based models (BERT as the example) and observe that 1) the attention matrices, $\mat{W_q}$ and $\mat{W_k}$ introduce much more gender bias than other modules (including the embedding layer) and  2) the bias degree changes periodically inside of the model (attention matrix $\mat{Q}$, $\mat{K}$, $\mat{V}$, and the remaining part of the attention layer (including the fully-connected layer, the residual connection, and the layer normalization module) enhance the gender bias while the averaged attentions reduces the bias). 
We hope our work will shine some lights on explainable and fairness AI.

\bibliographystyle{IEEEtran}
\bibliography{custom}

\end{document}